# Sketch2BIM: A Multi-Agent Human-AI Collaborative Pipeline to Convert Hand-Drawn Floor Plans to 3D BIM

Abir Khan Ratul[1], Sanjay Acharjee[2], Somin Park, Ph.D.[3], Md Nazmus Sakib, Ph.D.[4*]

[1] Ph.D. Student, Dept. of Civil Eng., University of Texas at Arlington. E-mail: abirkhan.ratul@uta.edu
[2] Ph.D. Student, Dept. of Civil Eng., University of Texas at Arlington. E-mail: sanjay.acharjee@uta.edu
[3] Assistant Professor, Dept. of Civil Eng., University of Texas at Arlington. E-mail: somin.park@uta.edu
[4] Assistant Professor, Dept. of Civil Eng., University of Texas at Arlington. E-mail: mdnazmus.sakib@uta.edu

## ABSTRACT

This study introduces a human-in-the-loop pipeline that converts unscaled, hand-drawn floor plan sketches into semantically consistent 3D BIM models. The workflow leverages multimodal large language models (MLLMs) within a multi-agent framework, combining perceptual extraction, human feedback, schema validation, and automated BIM scripting. Initially, sketches are iteratively refined into a structured JSON layout of walls, doors, and windows. Later, these layouts are transformed into executable scripts that generate 3D BIM models. Experiments on ten diverse floor plans demonstrate strong convergence: openings (doors, windows) are captured with high reliability in the initial pass, while wall detection begins around 83% and achieves near-perfect alignment after a few feedback iterations. Across all categories, precision, recall, and F1 scores remain above 0.83, and geometric errors (RMSE, MAE) progressively decrease to zero through feedback corrections. This study demonstrates how MLLM-driven multi-agent reasoning can make BIM creation accessible to both experts and non-experts using only freehand sketches.

---

[*] *Corresponding author*

## 1. INTRODUCTION

Building Information Modeling (BIM) has become a foundational technology in Architecture, Engineering, and Construction (AEC), offering rich, data-driven digital representations of buildings [1]. However, creating BIM models from scratch is still a resource-intensive process that demands significant expertise in specialized platforms such as Autodesk Revit [2]. This gap between early-stage conceptualization and digital modeling makes it difficult for beginners, non-BIM users, students, and those working in the early phases of design to adopt BIM effectively. Recent breakthroughs in deep learning, graph-based methods, and Large Language Models (LLMs) have advanced workflows for converting 2D floor plans into 3D BIM models [3-6]. While these approaches can effectively generate BIM models from scaled, grid-like floor plans, they fall short when dealing with unscaled, hand-drawn sketches that are irregular or abstract in nature. A key limitation is handling non-Manhattan layouts, where walls and spaces do not conform to the typical Manhattan-world assumptions of being strictly horizontal or vertical [7-9]. Instead, these layouts may feature angled, curved, or free-form geometries that deviate from orthogonal grids. This gap highlights the need for more adaptive and intelligent methods that can accurately handle such informal and irregular representations.

Hand sketches, in particular, remain central to the design process because they allow rapid and flexible exploration of ideas, reinforce spatial understanding, and capture the designer's conceptual intent before formalization in digital tools [10]. Prior studies emphasize that sketching not only enhances visual memory and perception of space but also strongly correlates with design ability, making it a critical pedagogical and professional practice [11, 12]. Thus, leveraging sketches as the primary input for automation ensures that the creative and cognitive foundations of early-stage design are preserved and faithfully translated into BIM environments.

Multimodal Large Language Models (MLLMs) offer a promising pathway forward [9]. These models can parallelly process and reason over multimodal data (e.g., visual, textual) inputs, producing structured outputs that are directly compatible with downstream applications [13]. Unlike unimodal approaches, that is used to process either images or text independently, MLLMs can interpret a hand-drawn floor plan in conjunction with accompanying textual descriptions, enabling a richer semantic understanding. This cross-modal reasoning capability makes MLLMs well suited for bridging unstructured inputs like freehand sketches with structured outputs such as detailed BIM models [2, 7]. Their ability to integrate domain knowledge, perception, and procedural reasoning in a single inference process reduces error propagation and supports end-to-end automation [14].

Building on that capability, this paper introduces a two-phase, human-in-the-loop multi-agent pipeline that converts a single hand-drawn floor-plan image into an auditable 3D BIM model. The first phase utilizes a vision-language model equipped with chain-of-thought prompting (Wei et al., 2022), visual grounding, and spatial reasoning to generate a schema-conformant 2D JSON representation of walls, doors, and windows, with each opening explicitly linked to its corresponding host wall. The output adheres to strict conventions, such as using feet for units, radians for arc angles, and counterclockwise rotation for polygons, to ensure that versions are directly comparable and traceable. Following the initial extraction, the system integrates human oversight by means of a web-based interface, allowing users to iteratively refine the extracted JSON layout through natural language interaction. The second phase uses a multi-agent controller that compiles the JSON into a deterministic script, which instantiates BIM geometry and metadata in Autodesk Revit. The multi-phased design deliberately separates perceptual inference from constructive synthesis. Phase 1 concentrates on uncertainty within a bounded, interpretable artifact (the JSON), whereas Phase 2 remains deterministic. This separation clarifies failure modes, simplifies debugging, and renders the end-to-end process auditable and reproducible.

By lowering the technical barrier to BIM creation, the proposed multi-agent pipeline has the potential to benefit professionals and non-professionals by facilitating faster prototyping, promoting consistent model standards, and expanding access to advanced modeling tools. In doing so, this approach may help bridge the gap between conceptual sketches and functional BIM models, moving toward a design-to-digital pipeline that is more efficient, interactive, and inclusive.

## 2. BACKGROUND STUDY

### 2.1. Foundational and Rule-based Methods

Early concepts in computational design, notably advanced by Negroponte and Eastman in the 1960s-1970s, laid the groundwork for algorithmic modeling in architecture [15, 16]. This vision matured in the 2000s with the introduction of parametric tools such as Generative Components (2003) and Grasshopper (2007), which brought rule-based workflows into mainstream architectural practice [7]. Eastman et al. (2009) formalized BIM-based rule checking into a staged pipeline of rule interpretation, model preparation, execution, and reporting, providing a systematic framework for model validation [17]. Later, Solihin and Eastman (2015) further classified BIM rule complexity to clarify semantic requirements for compliance-checking systems [18]. Hybrid workflows also emerged, such as Gimenez et al. (2015) combined automated processing with targeted user corrections to reconstruct 3D building models from scanned plans [19]. Collectively, these foundational approaches emphasized the need to balance automation with human oversight while integrating geometric, topological, and semantic reasoning.

## 2.2. Deep Learning Approaches

With the rise of data-driven methods, deep learning became central to floor plan interpretation. Liu et al. (2017) proposed a CNN-based approach using junction detection and integer programming to vectorize rasterized plans, effective for orthogonal layouts but limited to non-rectilinear geometries [4]. Jang et al. (2020) combined semantic segmentation with graph-based conversion to map floor plans into IndoorGML/CityGML standards [20]. Park and Kim (2021) developed 3DPlanNet, an ensemble CNN and rule-based pipeline that converted raster inputs into extruded 3D meshes [21]. More recently, Barreiro et al. (2023) reconstructed semantic 3D models from 2D plans using deep segmentation and vectorization pipelines [22]. These approaches have incredibly advanced automation of BIM, but still require clean, scaled, and often Manhattan-style inputs.

## 2.3. Generative and GAN-Based Approaches for 3D Form Generation

Researchers explored generative approaches to synthesize architectural forms. De Miguel Rodríguez et al. (2020) used variational autoencoders (VAEs) for interpolation-based form synthesis, though the method was constrained by limited training data diversity [3]. Yang et al. (2021) and Pouliou et al. (2023) extended this work through controllable point cloud GANs (CPCGANs), enabling rule-guided 3D point cloud generation but still lacking fine-grained architectural control [5, 23]. Kim et al. (2021) applied style-transfer GANs to normalize sketch variance before segmentation, improving accuracy on messy hand-drawn inputs [24]. While these methods introduced flexibility and generative design, they often struggled to maintain geometric fidelity and semantic consistency required for BIM.

## 2.4. LLM and Multi-Agent BIM Generation

The rise of Large Language Models (LLMs) has opened new possibilities for BIM automation. Du et al. (2024) introduced Text2BIM, a multi-agent framework that translates natural language descriptions into BIM code with iterative rule checking and live previews in Vectorworks [9].Building on this direction, Deng et al. (2025) leveraged multimodal LLMs to develop BIMgent, a hierarchical multi-agent system that simulates GUI interactions to construct BIM models [2]. BIMgent demonstrates an important step toward interactive BIM generation, but its reliance on pixel-level GUI grounding is fragile, and it does not provide robust semantic validation and natural language–based human-BIM interaction. Ratul et al. (2025) proposed a framework for automating the conversion of hand-drawn plans into 3D using MLLMs and 3D modeling software [7], demonstrating a promising direction toward multimodal integration. However, the work focuses only on simple rectangular or triangular floor plans and lacks mechanisms to edit or refine outputs when inaccuracies occur. These developments underscore the potential of LLM-driven orchestration but highlight the need for approaches that better integrate perception, reasoning, and user feedback.

## 2.5. Gap and Motivation for This Work

Despite significant progress, the automated transformation of unscaled, hand-drawn sketches into semantically consistent BIM models remains underexplored. Existing deep learning methods assume clean and rectilinear inputs, while generative and GAN-based approaches often lack semantic fidelity. Recent LLM-based systems such as Text2BIM and BIMgent highlight the potential of multi-agent orchestration but remain fragile in grounding and limited in interactive validation. To address these gaps, this study introduces a feedback-driven multi-agent pipeline that combines multimodal LLM's perception, symbolic reasoning, schema enforcement, and natural language feedback. This design enables robust, tunable BIM creation from raw sketches, bridging geometric fidelity with semantic correctness while keeping humans feedback in the loop.

## 3. PROBLEM STATEMENT AND RESEARCH OBJECTIVES

### 3.1. Problem Statement

Despite decades of progress in computer vision, deep learning, and generative AI, the automated interpretation of floor plans into structured 3D BIM remains an unsolved challenge. Early approaches, grounded in rule-based or image-processing techniques, could extract primitives such as walls and openings but struggled with noisy, irregular, sophisticated and diverse drawings, often requiring semi-automatic post-processing. Advancements in deep learning, including CNN-based segmentation, vectorization via integer programming, and generative methods such as VAEs and GANs, have improved accuracy but remain heavily reliant on large, curated datasets and fail to generalize to unscaled, hand-drawn sketches.

More recent agentic systems demonstrate the promise of orchestrating perception and reasoning. However, these frameworks remain predominantly one-shot, fully automatic pipelines that assume structured, scaled, and grid-aligned inputs. They lack robust mechanisms to handle non-Manhattan drawings with angled walls or curved geometries, and they provide no structured human-in-the-loop feedback during the critical extraction Phase. Furthermore, their reliance on brittle pixel-level grounding and rigid API toolsets limits semantic consistency, adaptability, and parametric editability.

This research addresses these limitations by introducing a feedback-driven, multimodal, multi-agent pipeline designed explicitly for unscaled, hand-drawn, and non-Manhattan sketches. The proposed pipeline combines MLLM reasoning, geometric extraction, and human-guided validation to refine noisy, informal inputs prior to generating BIM models. By unifying automated perception with structured feedback, the proposed system enhances accuracy, adaptability, and usability across diverse architectural workflows.

## 3.2. Research Objective and Contributions

The objective of this research is to develop a hybrid automated–interactive pipeline for converting hand-drawn floor plans into semantically consistent BIM models. The main contributions are:

1. Developing a framework to process unscaled hand-drawn floor plan sketches and move beyond reliance on CAD-like inputs or large annotated datasets.

2. Combining MLLM reasoning with geometric extraction to capture both Manhattan and non-Manhattan layouts, including angled and curved walls.

3. Embedding a structured human feedback stage in 2D coordinate extraction, increasing reliability and overcoming the weakness of one-shot automated method.

4. Orchestrating perception, validation, and synthesis in a modular pipeline, reducing manual modeling effort while maintaining semantics and industry-standard accuracy.

## 4. METHODOLOGY

This study proposes a human-in-the-loop multi-agent pipeline that transforms unscaled 2D floor plan sketches into a 3D BIM model. The pipeline comprises:

1. Phase 1: Layout Extraction

2. Phase 2: Revit Synthesis

Across both phases, OpenAI's MLLM GPT-5 was used as the core agent for multimodal perception, geometric reasoning, and code synthesis, ensuring accurate extraction and reliable BIM instantiation. GPT-5 was chosen for its multimodal capacity to handle both text and images, its strong reasoning for complex spatial layouts, and its ability to generate executable code, making it well-suited for transforming unstructured sketches into a structured 3D BIM model [25].

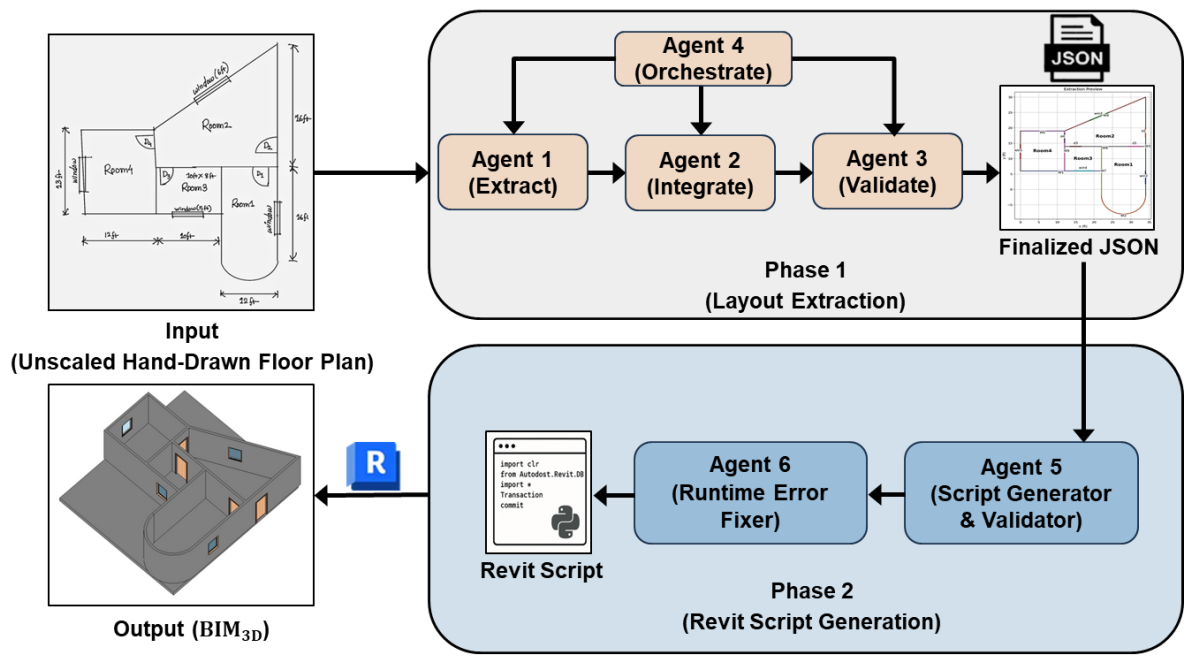

**Fig. 1.** Overview of the workflow for agentic BIM creation from hand-drawn floor plans.

Fig. 1 illustrates the overview of the end-to-end pipeline that integrates layout extraction with Revit BIM generation. The process begins with a hand-drawn, annotated floor plan image [ $I \in R^{H \times W \times 3}$; $represent\ a\ three-channel\ image\ with\ height\ H\ and\ width\ W$ ] as input, processed by a multi-agent system comprising four specialized agents responsible for perceptual extraction, human-feedback refinement, schema validation, and orchestration. The validated layout is then translated into executable Python code for RevitPythonShell (RPS) by a second multi-agent system focused on code generation and validation. The final $BIM_{3D}$ output is executed within Revit.

## 4.1. Phase 1: Layout Extraction (Sketch→JSON)

The first phase of the pipeline transforms a raw, hand-drawn, unscaled raster floor plan ($I \in R^{H \times W \times 3}$) into a schema-conformant JSON layout ($J_i$) through a coordinated, human-in-the-loop, multi-agent system. Phase 1 includes a total of four agents (i.e., $A_1$ to $A_4$) and each agent employs a structured chain of thought, prompting internal planning to perceive, apply geometric rules, and route fixes. Dimension callouts and labels (e.g., Door/D, Window/W/Win, Room) provide weak priors but are unreliable due to scale ambiguity and sketch imperfections.

Agent $A_1$ (Perceptual Extraction, GPT-5) processes the base64-encoded image using chain-of-thought prompting to jointly interpret visual geometry, line structure, and textual annotations. The output is an initial structured layout ($J_0$):

$$J_0 = \{e_k\}_{k=1}^{N}, \quad e_k = (c_k, x_k, p_k) \tag{1}$$

Where, $e_k$ is an element of the floor plan, $c_k \in C$ is the element class (e.g., wall, door, window), $x_k \in R^d$ are 2D coordinates, and $p_k$ are auxiliary attributes (e.g., thickness, material, orientation). Alongside $J_0$, the agent produces a semantic summary ($S$) describing topology, element counts, adjacency relations, and approximate proportions, allowing later iterations to reuse contextual reasoning without re-encoding the input image ($I$).

Agent $A_1$ starts with preprocessing to ensure that global distortions are removed without forcing Manhattan alignment. Hough peaks function [26] is used to estimate a global skew angle $\varphi$, which is subtracted from all detected lines. Pixel-to-foot scaling is computed from annotated dimensions using anisotropic least-squares with regularization toward isotropy, giving consistent scaling before walls and rooms are instantiated.

Agent $A_1$ treats Walls in the floor plan as either straight, angled, or curved. Straight walls are first detected using probabilistic Hough methods [26] together with Random Sample Consensus (RANSAC) [27] and then divided wherever they meet at intersections or T-junctions. After detection, the system groups wall orientations into clusters. The number of these clusters, $K \in [1, 8]$, is chosen automatically using two statistical criteria, the Bayesian Information Criterion (BIC) and the Akaike Information Criterion (AIC) [28]. These measures help the system determine the number of distinct wall directions that actually exist in the sketch. By doing this, the method avoids forcing all walls to be aligned only in vertical and horizontal directions, meaning it can correctly handle drawings that include diagonals or other non-rectangular layouts. Local merging occurs only when the orientation difference is $\leq 1°$, endpoints are within 3.0 ft, perpendicular offset is $\leq 0.50$ ft, and overlap is $\geq 60\%$. Double-wall strokes that run parallel with a consistent gap are merged into a single wall centerline. When a segment changes direction by more than 5 degrees from the dominant wall orientation, it is treated as part of a new angled wall cluster, ensuring that non-orthogonal geometries are preserved. Short, angled wall stubs are not discarded automatically; instead, they are retained as structural elements if they are at least 1.25 feet long, connect two longer walls forming a non-180° turning angle of at least 12°, and contribute to closing a bounded room shape such as a trapezoid, wedge, or other irregular enclosure. This ensures that even small connecting walls, important for defining room boundaries, are preserved rather than removed during simplification. Curved walls are represented as arcs whenever the radius, center, and angles are available; if missing, they are computed using the sagitta midpoint method (arc3pt). Chord endpoints are always included and the sweep angle ($\Delta\theta$) is normalized to $0 < \Delta\theta < 2\pi$. The sagitta error is limited to 0.10 ft.

Agent $A_1$ uses available text labels or swing arcs from the sketch to detect openings such as doors and windows. The host wall is assigned by nearest adjacency and tangent consistency. For line walls, openings are projected orthogonally; for arc walls, they are projected as tangent chords. Default sizes are used if labels are missing. For example, doors are typically around 3.0 ft in width (ranging from 2.5 ft to 3.5 ft), and windows are about 4.5 ft in width (ranging from 4.0 ft to 5.0 ft). Openings must be placed at least 0.75 ft from wall ends and no closer than 0.50 ft to each other; they cannot overlap or cross wall vertices. If violations occur, openings are trimmed or relocated. Rooms are extracted by building a half-edge planar graph from the final walls. Openings are annotated on walls but remain uncut. Bounded faces are enumerated, the exterior face is removed, and the remaining polygons are output as rooms. Concavities and non-Manhattan shapes are preserved by following the actual wall chain counterclockwise. Sliver polygons [29] narrower than 2.0 ft are merged into neighboring rooms. All room polygons are closed, counterclockwise, and non-self-intersecting.

The second agent $A_2$ (Human Feedback Integration, GPT-5) refines layouts $J_i$ based on user feedback in natural language ($F_i \in$ Accept, Reject, Edit). The update rule is

$$J_{i+1} = \begin{cases} J_i & F_i = Accept \\ A_2(J_i,\ F_i,\ S_i) & Otherwise \end{cases} \quad (2)$$

Here, $S_i$ is the system state, containing the sketch features, feedback history, geometric constraints, and schema rules that guide the layout update. The agent $A_2$ supports user-directed edits, including merging or splitting walls, adding or moving doors and windows, correcting IDs, and adjusting wall thickness. For consistent and deterministic ID assignments, outer boundary walls are numbered first, interior walls are then ordered from left to right and bottom to top, openings are grouped under their respective host walls, and rooms follow the same ordering convention.

The third agent $A_3$ (Schema and Topology Validation, GPT-5) checks that outputs satisfy both syntax and geometry. Validation is further enforced by a strict JSON checker. It ensures schema compliance, unique ID, and geometric consistency. Walls must have nonzero length or radius, arcs must have sweep $< 2\pi$, and arc3pt definitions must not be collinear. Openings must have consistent length, lie on walls and remain at least $0.75$ ft from wall ends. Rooms must be counterclockwise, non-self-intersecting, and have positive area.

The validator reports issues along with suggested fixes. Typical detections include syntax errors, duplicate IDs, and disallowed full-circle arcs. It also identifies and refits nearly collinear arc3pt definitions, realigns openings that are misaligned with their host walls, trims openings intersecting wall corners, and separates or merges overlapping openings. Additional checks address missing wall references by attaching them to the nearest wall, rooms that are not properly closed or snapped to walls, and coordinates with excessive decimal precision, these are rounded for consistency. The validation indicator is

$$V(J) = \left[ schema(J) \wedge closed\_polygons(rooms) \wedge openings \subseteq walls \wedge \forall w \right. \tag{3}$$
$$\left. \in walls,; connected\left(p_{\{start\}(w)}\right) \wedge connected\left(p_{\{end\}(w)}\right) \right]$$

Where $connected(p)$ is true if endpoint $p$ coincides with another element or lies on a valid curve. If $V(J) = 0$, a violation set $C_i$ is returned to guide repair. The fourth agent $A_4$ (Orchestration Controller, GPT-5) supervises the process, maintaining memory,

$$M_i = (I_{id}, S, J_i, H_i, C_i) \tag{4}$$

Where $I_{id}$ is the input identifier, $S$ the semantic summary, $J_i$ the current layout, $H_i$ the cumulative human feedback, and $C_i$ the set of detected constraints.

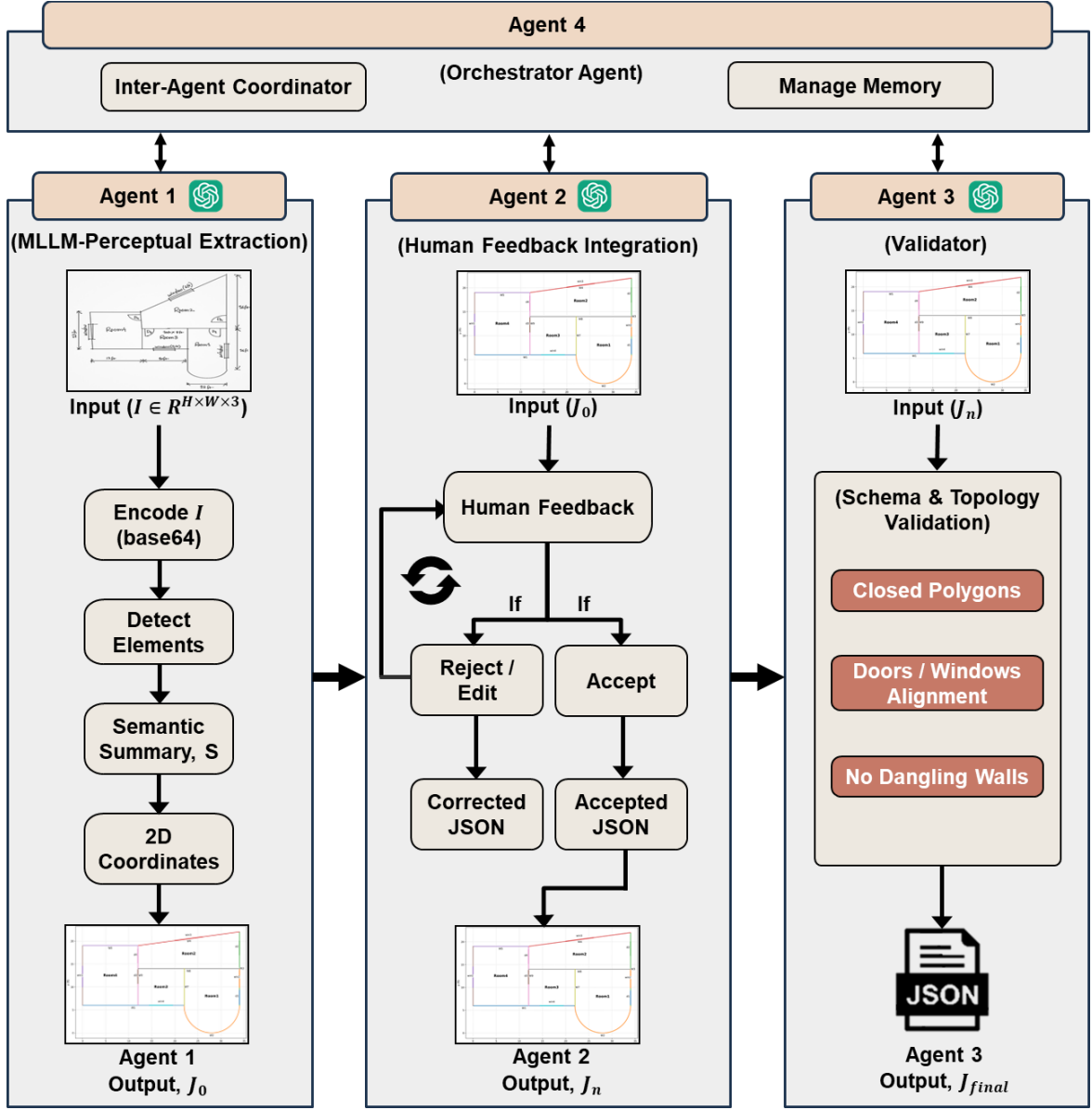

**Fig. 2.** Workflow for Phase 1 (layout extraction from sketch).

The refinement policy is,

$$\pi_1(J_i, F_i, C_i) = \begin{cases} halt, \ J_{final} = J_i, & F_i = Acceppt, V(J_i) = 1 \\ A_2(J_i, F_i, C_i), & F_i \in Reject, Edit, V(J_i) = 1 \\ A_2(A_3(J_i), F_i, C_i), & V(J_i) = 0 \end{cases} \quad (5)$$

The process halts only when both human accepts, and validation passes. Together, perceptual extraction, human feedback, validation, and orchestration ensure that $J_{final}$ is topologically sound and semantically consistent, robust to straight, angled, and curved walls, as well as doors, windows, and complex non-Manhattan geometries. Fig. 2 shows the workflow of the first phase of the pipeline.

## 4.2. Phase 2: Revit Script Generation and Validation

The finalized structured layout $J_{\text{final}}$ from the first phase is transformed into an executable Python script ($P$) compatible with the Autodesk Revit API. Autodesk Revit was selected because it is one of the most widely used BIM platforms in AEC. This process ensures that geometric, semantic, and procedural constraints are respected. Two agents are employed in this phase: $A_5$ for script synthesis and static validation, and $A_6$ for runtime error correction. Both rely on GPT-5 for high-reasoning and code generation, while feedback from RPS execution, a scripting environment for automating tasks in Autodesk Revit, is used iteratively. The initial mapping is defined as:

$$P_0 = A_5(J_{\text{final}}) \quad (6)$$

Where $A_5$ translates each element $e_k \in J_{\text{final}}$ into Revit API calls:

$$e_k \mapsto \text{call}_{\text{Revit}}(c_k, x_k, p_k) \quad (7)$$

Ensuring that objects such as Wall, Family Instance, and Floor are instantiated in a dependency-consistent order. To detect inconsistencies before execution, $A_5$ invokes a validation module $\Gamma_{\text{script}}$ (script_guard.py) that applies static semantic checks:

$$V_P(P) = [\, imports(P) \wedge syntax(P) \wedge api\_bindings(P) \wedge dependency\_order(P) \,] \qquad (8)$$

Where $V_P(P) \in 0,1$ indicates whether $P$ passes all structural code constraints. If $V_P(P) = 0$, the set of violations $C_P$ is returned. The runtime error-correction agent $A_6$ operates in a closed loop with RPS execution. Given runtime error traces $E_t$, it applies targeted code edits:

$$P_{t+1} = A_6(P_t, E_t) \qquad (9)$$

This allows $A_6$ to resolve both Revit API exceptions and logical execution faults, using error semantics and the original $J_{\text{final}}$ as context.

$$\Pi_2(P_t, C_P, E_t) = \begin{cases} halt, P_{final} = P_t & V(P_t) = 1 \wedge E_t = \emptyset \\ A_6(P_t, E_t), & V(P_t) = 1 \wedge E_t \neq \emptyset \\ A'_5(J_{final}, C_P), & V(P_t) = 1 = 0 \end{cases} \qquad (10)$$

Where $A'_5$ denotes the repair mode of $A_5$, regenerating corrected code based on static validation failures $C_P$, This phase terminates once produces a procedurally accurate and semantically consistent 3D BIM model within Revit:

$$P_{final} \xrightarrow{\text{RevitPythonShell}} BIM_{3D} \qquad (11)$$

Fig. 3 summarizes phase 2 of the pipeline. Here, the validated JSON layout $J_{\text{final}}$ is converted into an executable script. The Script Generator ($A_5$) maps layout entities to Revit API calls and applies static validation rules. The Error-Correction Agent ($A_6$) iteratively repairs scripts based on runtime traces until both validation and execution succeed. The final $P_{\text{final}}$ I executed in Revit through RPS to generate the $BIM_{3D}$ model.

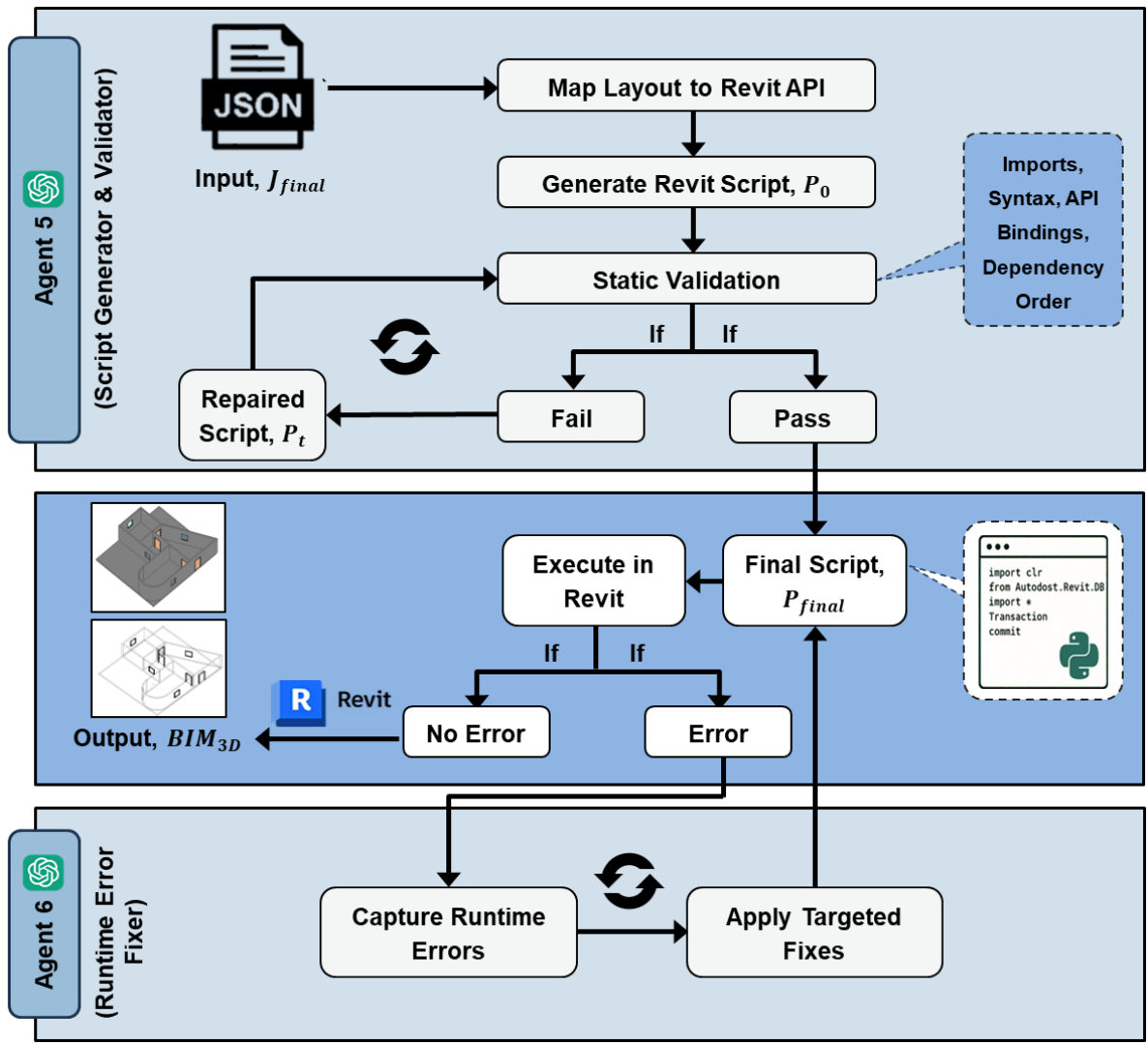

**Fig. 3.** Flowchart of Phase 2 (Revit script generation).

## 5. EVALUATION AND PERFORMANCE ANALYSIS

This study evaluates the pipeline, where Phase 1 converts a hand-drawn floor plan into a structured 2D JSON file containing walls (represented by straight lines and circular arcs), doors, and windows, all anchored to host walls. Phase 2 then synthesizes this JSON file into a 3D BIM model in Autodesk Revit. The analysis concentrates on phase 1 because errors at extraction directly propagate into BIM. The evaluation process investigated (1) whether the correct set of elements was found and (2) if detected, how accurate the recovered geometry is. Scoring is performed against a human-curated ground truth (GT). After each round of human feedback, the predictions are re-scored to measure improvements at the overall and at each edit. All tolerances and penalties are derived automatically from GT statistics so that errors scale with plan size and typical element lengths.

## 5.1. Pairing and Matching Policy

Predicted elements are paired to GT strictly within category walls to walls, doors to doors, and windows to windows to avoid cross-type matches. Pairing proceeds ID-first after normalizing case and spacing, to mirror how edits are logged; for openings, numeric suffixes must agree so that "$win11$" never pairs with "$win15$" and "d2" pairs with "d02" only if both normalize to "2." When IDs do not resolve a unique match, pairing falls back to optimal assignment under geometric gates. Concretely, a category-specific cost matrix is built from mid-point distances, length similarity, and, for lines, small orientation differences; arc matches are additionally gated by radius consistency. The final one-to-one pairing is obtained with the Hungarian assignment method on a square-extended cost matrix, which respects hard gates that disallow implausible matches. Openings are host-aware: the wall pairing is used to map predicted opening hosts into GT wall indices; when a host is unknown, nearest-wall logic by projected midpoint is used. This design follows standard practice in optimal assignment and robust model fitting [30, 31] to detect straight segments using probabilistic Hough and prune outliers with RANSAC before forming costs, then solve the bipartite assignment via the Hungarian method.

## 5.2. Metrics for Detection Quality

Set quality is quantified using precision, recall, and their harmonic mean $F1$, reported per category (walls, doors, windows) and as a micro-averaged overall score. True positives ($TP$) are correctly matched predictions; false positives ($FP$) are unmatched predictions; false negatives ($FN$) are unmatched ground-truth instances. The metrics are defined as

$$\text{Precision} = \frac{TP}{TP + FP} \tag{12.1}$$

$$\text{Recall} = \frac{TP}{TP + FN} \tag{12.2}$$

$$F1 = \frac{2 \cdot \text{Precision} \cdot \text{Recall}}{\text{Precision} + \text{Recall}} \tag{12.3}$$

Matching is one-to-one: IDs are used first when available, followed by Hungarian assignment under geometric gates (orientation and proximity), to ensure consistent pairing. Evaluation is reported per category and as micro-averaged overall scores. Precision captures the absence of spurious elements, while recall reflects the recovery of true ones. Unlike conventional classification tasks, there are no true negatives in this setting since elements do not present in the scene are never detected, which makes the accuracy metric less meaningful. Instead, $F1$ is used because it balances precision and recall, providing a more reliable measure of detection performance under class imbalance.

### 5.3. Metrics for Geometric Fidelity

After detection, evaluation must verify not only whether the correct elements were found but also whether their dimensions and placements are accurate. A model can score well on precision/recall yet still output drawings that are mis-sized or misaligned, and this breaks topology, distorts areas, and increases post-editing. Geometric fidelity, therefore, focuses on matched (TP) pairs and measures two complementary aspects: length accuracy and positional accuracy. For walls, length refers to the line segment length (or arc length, if curved); for openings, length is the span along the host wall. The per-pair length error ($\Delta L_i$) is defined as the difference between the predicted ($L_i^{\text{pred}}$) and GT ($L_i^{\text{GT}}$) length ($\Delta L_i = L_i^{\text{pred}} - L_i^{\text{GT}}$). Length accuracy is aggregated with Root Mean Square Error (RMSE), which emphasizes larger deviations that materially impact dimensions and downstream analyses:

$$RMSE = \sqrt{\frac{1}{N}\sum_{i=1}^{N}(\Delta L_i)^2} \tag{13}$$

Positional discrepancy captures spatial misalignment even when sizes are correct. It is measured as the Euclidean distance between the midpoints of the paired elements ($\Delta p_i = ||m_i^{\text{pred}} - m_i^{\text{GT}}||_2$). Positional accuracy is summarized using the Mean Absolute Error (MAE), which reflects the average shift without over-weighting rare extremes.

$$MAE = \frac{1}{N}\sum_{i=1}^{N}|\Delta p_i| \tag{14}$$

These two metrics are required because they expose distinct, practically relevant failure modes [32]. RMSE on length answers how far an element's size deviates, penalizing large over/under-sizing that propagates into room dimensions, areas, and performance estimates. MAE on position answers how far an element is displaced, revealing drifts that break alignments, thresholds, and circulation, and that directly increase editing burden. Reporting both per category (walls, doors, and windows) ties evaluation to absolute BIM tolerances and structural continuity for walls and edit load and usability for openings, providing a complete picture of geometric quality beyond detection alone.

## 5.4. Thresholds, Scaling, and Iterative Scoring

All matching and validation thresholds are derived automatically from GT statistics so that they remain proportional to the overall plan scale. For example, line-to-line orientation gates are set tightly to prevent incorrect pairing of non-parallel walls, while arc matching requires consistent radii to ensure curved segments are not confused. Positional gates are scaled by typical element lengths so that only spatially plausible pairs are considered, preventing distant items from being matched. Additional numeric tolerances used throughout the pipeline, such as the embedding tolerance for openings within walls, the minimum spacing between adjacent openings, and the rejection of extremely narrow "sliver" polygons, are specified in feet and chosen to capture realistic spacing relationships in architectural layouts. This scaling strategy avoids over-penalizing minor sketching imperfections while ensuring that deviations large enough to affect BIM usability are flagged.

The scoring process is iterative. At iteration $i$, the extraction produces a layout $J_i$. Human feedback $F_i$ is then incorporated to update the layout into $J_{i+1}$ After this pairing, all metrics are recomputed. In practice, early edits such as restoring a missed wall, relocating a misplaced opening, or correcting a host-wall assignment primarily improve recall by increasing true positives and thereby reducing false negatives. Subsequent edits refine geometry, leading to reductions in positional and length errors. Because the integrated metric penalizes both false positives and false negatives, it is sensitive to these improvements on both fronts and serves as a compact objective function to track convergence toward the GT layout.

## 5.5. Validation of Phase 2

After script synthesis and execution in RPS, the resulting $BIM_{3D}$ is validated by human reviewers to ensure that Phase 1 quality translates into a correct, editable model. Reviewers verify that JSON walls, doors, and windows are mapped to appropriate Revit classes, that wall connectivity is continuous without gaps or dangling segments, that openings are embedded on their hosts with correct offsets and spans, and that angled and curved geometries are preserved without Manhattan simplification. This qualitative check confirms that any residual issues not captured by Phase 1 metrics, such as API binding errors or instantiation quirks, are resolved before downstream use. Human validation involved three graduate students in BIM with advanced training and research experience, who reviewed all ten floor plans.

## 6. CASE STUDIES AND DATA

The proposed system was evaluated on a set of ten hand-drawn floor plans created in the Samsung Notes app on a Samsung Tab S10+ using the stylus pen, a setup chosen to mimic natural sketching conditions encountered in early design phases closely. To encompass diverse and challenging scenarios, these drawings were deliberately designed to combine Manhattan and non-Manhattan geometries, concavities, skew, and curved walls, as summarized in Table 1. Such conditions introduce three sources of failure in automated extraction: (1) box-bias, where algorithms incorrectly "orthogonalize" building exteriors; (2) parallel-wall ambiguity, where openings are mistakenly matched to the wrong host wall; and (3) small-edge volatility, where short stubs or micro-edges either fragment structural boundaries (false positives) or are suppressed altogether (false negatives). These challenges provide a robust testbed for assessing the advantages of our geometry-gated, host-aware matching and the coverage-weighted loss formulation. In addition, the dataset captures a range of drawing complexities, including irregular layouts and variable stroke thicknesses that resemble real user behavior. This diversity ensures that the evaluation not only benchmarks geometric accuracy but also tests the system's resilience to noisy, imperfect sketches that are typical in conceptual architectural design.

**Table 1.** Floor plan set and evaluation focus.

| Plan | Class | Summary | Curved | Skew | Primary focus |
|------|-------|---------|--------|------|---------------|
| P01 | Manhattan | 3 rooms: rectangular | No | No | Baseline rectilinear extraction; opening placement |
| P02 | Manhattan | 5 rooms: rectangular | No | No | Orientation robustness; window sizing; host assignment on parallel |
| P03 | Manhattan | 5 rooms: square, rectangle, octagon | No | Yes | Regular polygon handling; stability with many short edges |
| P04 | Manhattan | 3 rooms: octagon, rectangle | No | No | Multi-edge polygons; length stability across joints |
| P05 | Manhattan | 5 rooms; rectangle; global skew | No | Yes | Deskew without forcing 0/90°; exterior from wall strokes only |
| P06 | Manhattan | 7 rooms: rectangular, trapezoidal | No | Yes | Preserve trapezoids; correct merge/split at T-junctions |
| P07 | Non-Manhattan | 3 rooms: trapezoidal, triangular, curved | Yes | No | Arc extraction; openings as tangent chords; host-aware matching |
| P08 | Non-Manhattan | 4 rooms: rectangular, trapezoidal, curved | Yes | No | Orientation clustering beyond 0/90°; line–arc intersections |
| P09 | Non-Manhattan | 5 rooms: rectangular, trapezoidal, triangular, curved | Yes | Yes | Heterogeneous shapes; coverage vs. numeric fidelity |
| P10 | Non-Manhattan | 5 rooms: curved, trapezoidal, polygon | Yes | Yes | Full stress test: concavities + arcs |

Fig. 4 provides a detailed example of the extraction process for floor plan 10, a layout comprising five rooms with curved and trapezoidal geometries and polygonal boundaries. The pipeline begins with the initial automated interpretation of the raw sketch, followed by iterative refinement through human-in-the-loop feedback. In total, eight user feedback steps were required to finalize the layout.

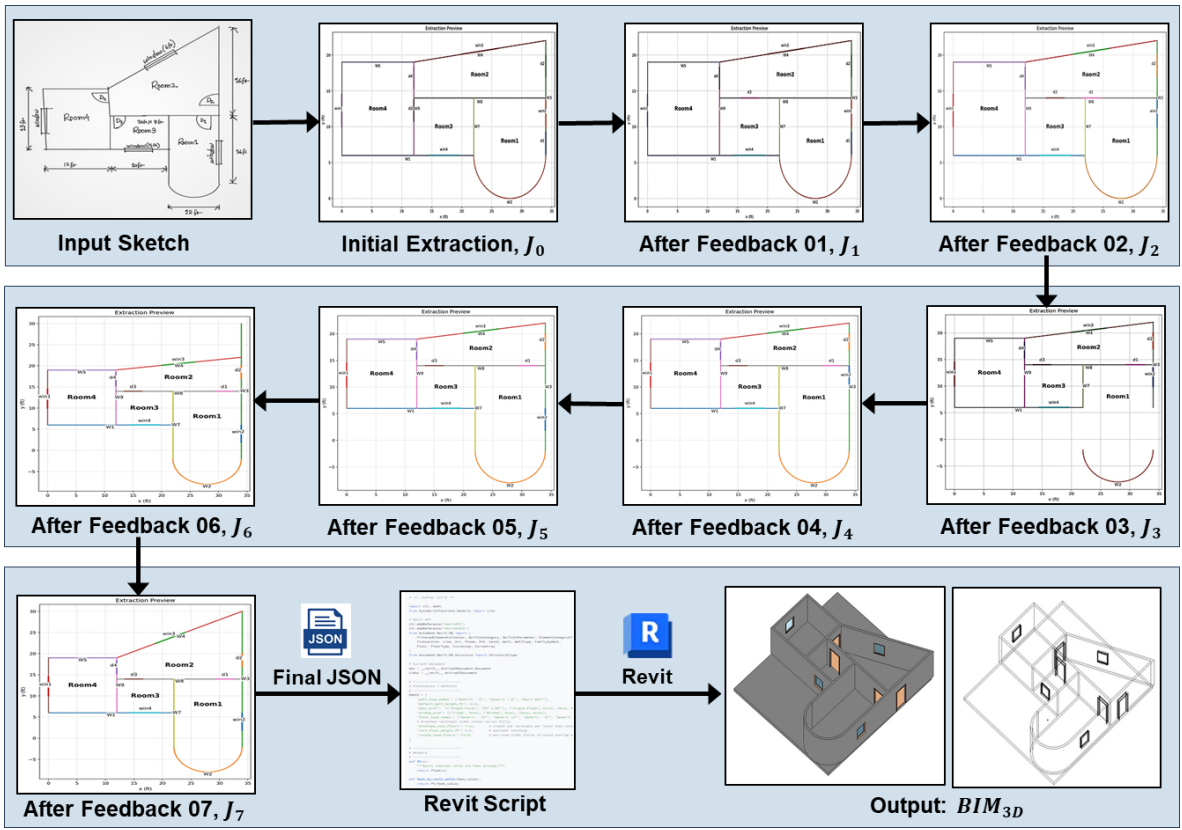

**Fig. 4.** Extraction workflow for floor plan 08, from sketch to final 3D BIM in Revit.

**Table 2.** User feedback for floor plan 10.

| Step | Action type | Target element(s) | User feedback (F) |
|------|-------------|-------------------|-------------------|
| 1 | Move | Door 3 | Move door 3 to wall 8 |
| 2 | Move | Door 1 | Move door 1 to wall 8 |
| 3 | Move/Adjust | Door 1, Door 3, Wall 2 | Move door 1 eight feet to the right, move door 3 two feet to the left, and move wall 2 downward by 8 feet |
| 4 | Extend | Wall 7, Wall 3 | Extend wall 7 and wall 3 downward by 8 feet |
| 5 | Move | Window 2 | Move window 2 downward by 8 feet |
| 6 | Extend | Wall 3 | Extend wall 3 upward by 8 feet |
| 7 | Connect | Wall 4, Wall 3, Wall 5 | Connect wall 4 with the top end of wall 3, while keeping the left end of wall 4 connected to wall 5 |

Table 2 summarizes these feedback steps, each expressed in natural language, specifying the action type (e.g., move, remove, shorten, connect, place) and the target element (walls, windows, or doors). These corrections progressively adjusted the initial extraction by shifting, resizing, and reassigning elements until the 2D layout was accurate enough to be finalized and stored as a JSON file. The finalized JSON was then translated into a Revit script and executed in RPS, producing the accurate 3D BIM model in Revit.

To enhance the usability of the system, two integrated web applications were developed using Streamlit, an open-source Python framework for building interactive web apps. The first application enables users to iteratively refine the 2D extraction through natural language feedback, producing a finalized JSON representation of the floor plan. The second application converts this JSON into a runnable Revit script that can then be executed within Revit to generate the BIM model. Fig. 5 illustrates the interfaces of these applications.

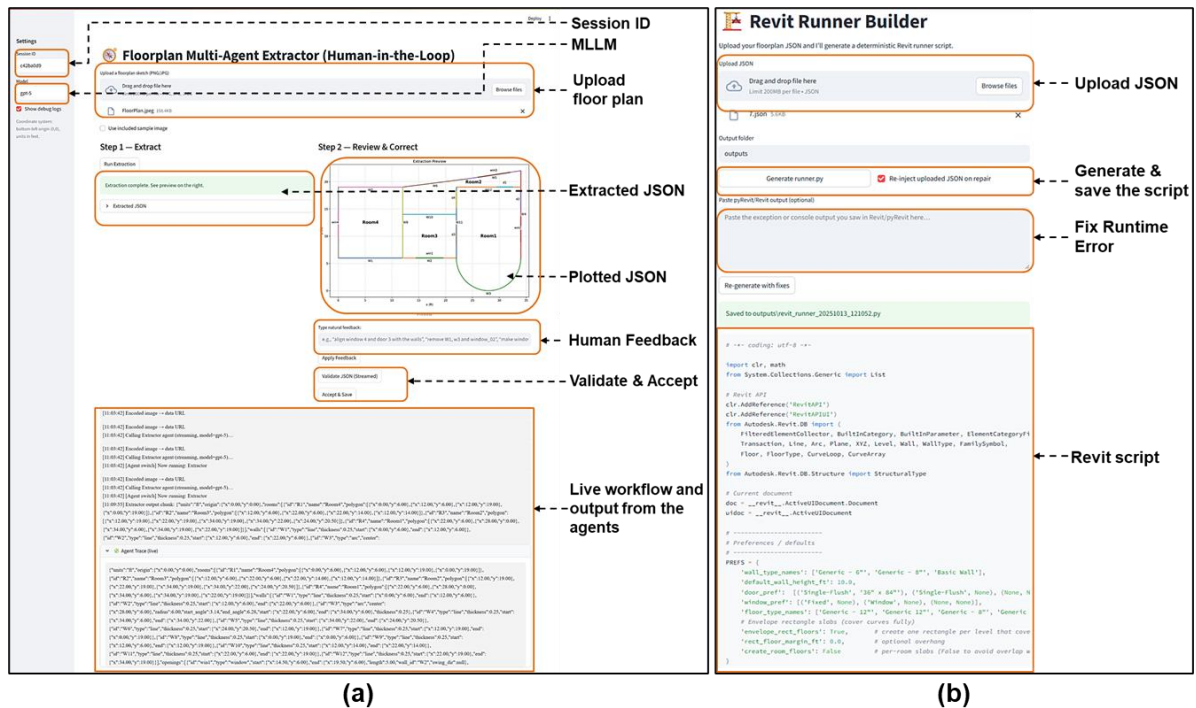

**Fig. 5.** Web applications for end users. (a) Streamlit app for iteratively refining the 2D extraction. (b) Streamlit app for converting the finalized JSON into a Revit script.

# 7. RESULTS

## 7.1. Aggregate Detection Performance (Phase 1)

The detection performance of the proposed pipeline was evaluated across ten floor plans, with a focus on the iterative refinement of walls, doors, windows, and overall layouts. Fig. 6 summarizes the convergence behavior using $\Delta F1$ heatmaps, while Figs. 7, 8, and 9 plot per-plan trajectories of precision, recall, and $F1$ score across iterations.

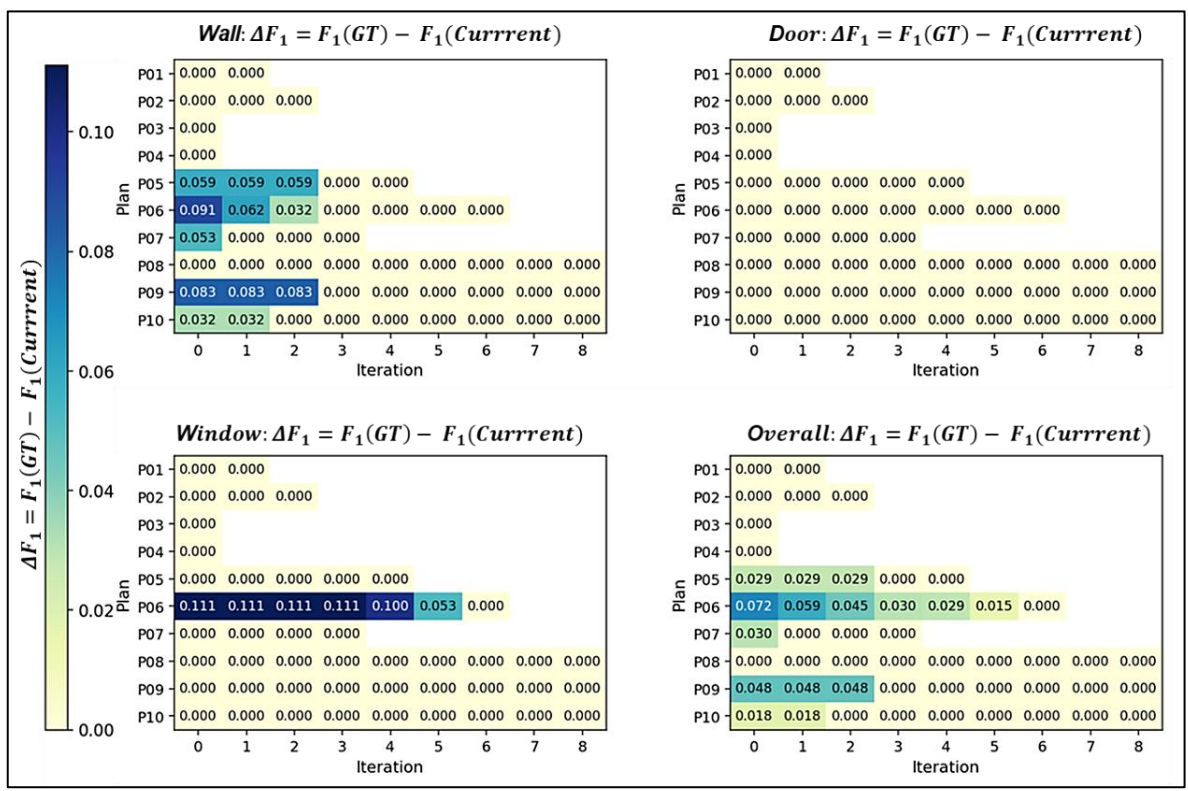

**Fig. 6.** $\Delta F1$ Heatmaps across plans and iterations for walls, doors, windows, and overall categories, showing convergence of detection performance toward GT.

The $\Delta F1$ heatmaps in Fig. 6 highlight how rapidly extracted layouts approach their final accuracy ($F1 = 1.0$). Wall detection stabilizes almost immediately for simpler floor plans, and for complex floor plans, the maximum $\Delta F1$ recorded was 0.091. For all floor plans, all the Doors were detected accurately. In contrast, windows detection showed a decline, with a $\Delta F1 = 0.11$ observed for one floor plan (P06). Despite this, in all cases the layout accurately converges within three to four iterations, emphasizing the effectiveness of the human-in-the-loop corrections. The overall category closely parallels the behavior of walls, suggesting that structural boundaries dominate aggregate performance.

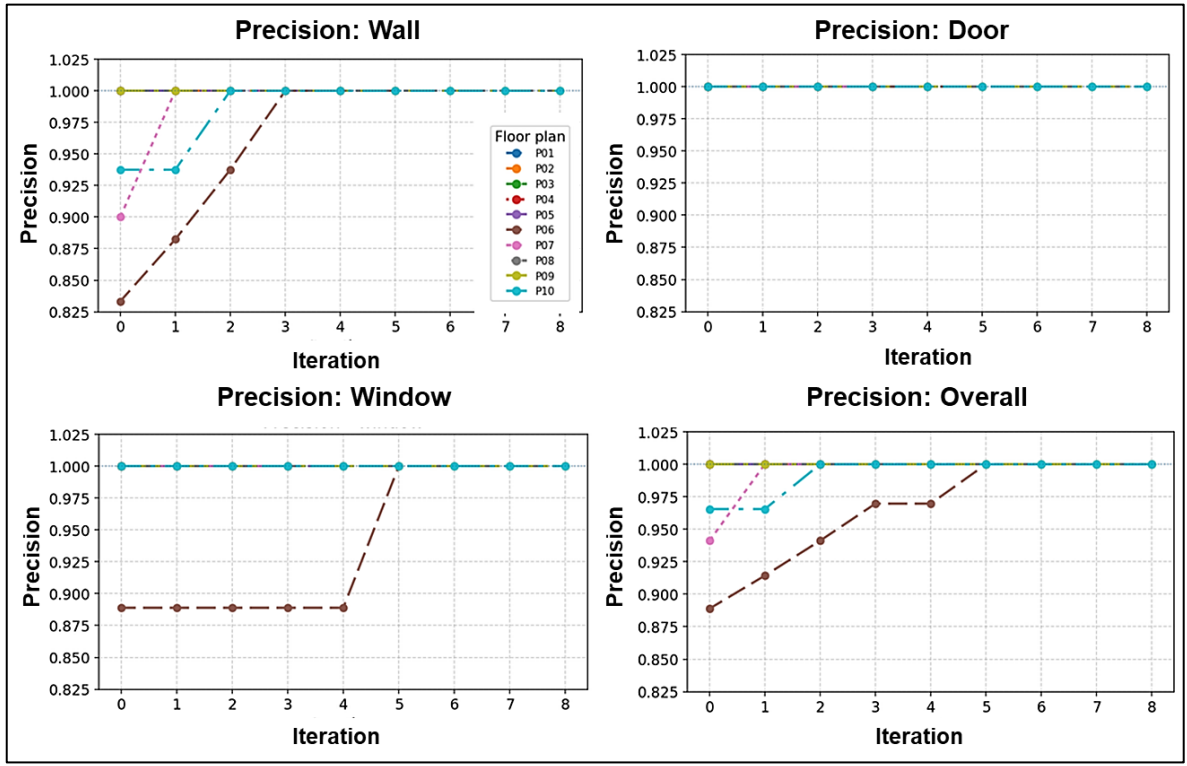

**Fig. 7.** Precision vs. human-feedback iteration by floor plan (P01–P10) and element.

Fig. 7, shows the precision trajectories for each plan and element. Precision increases rapidly and typically reaches 1.0 within the first iteration for walls, indicating that false positives are minimal from the start. Doors achieve a precision of 1.0 in the initial extraction, reflecting the model's strong ability to correctly identify openings without generating spurious detections. Windows show very limited variation across plans, with only minor deviations observed in one case. The overall precision curve closely follows that of the walls, confirming that structural elements largely determine the aggregate detection trend.

The overlapping dashed lines at 1.0 illustrate that most floor plans converge to perfect precision after only minimal user feedback.

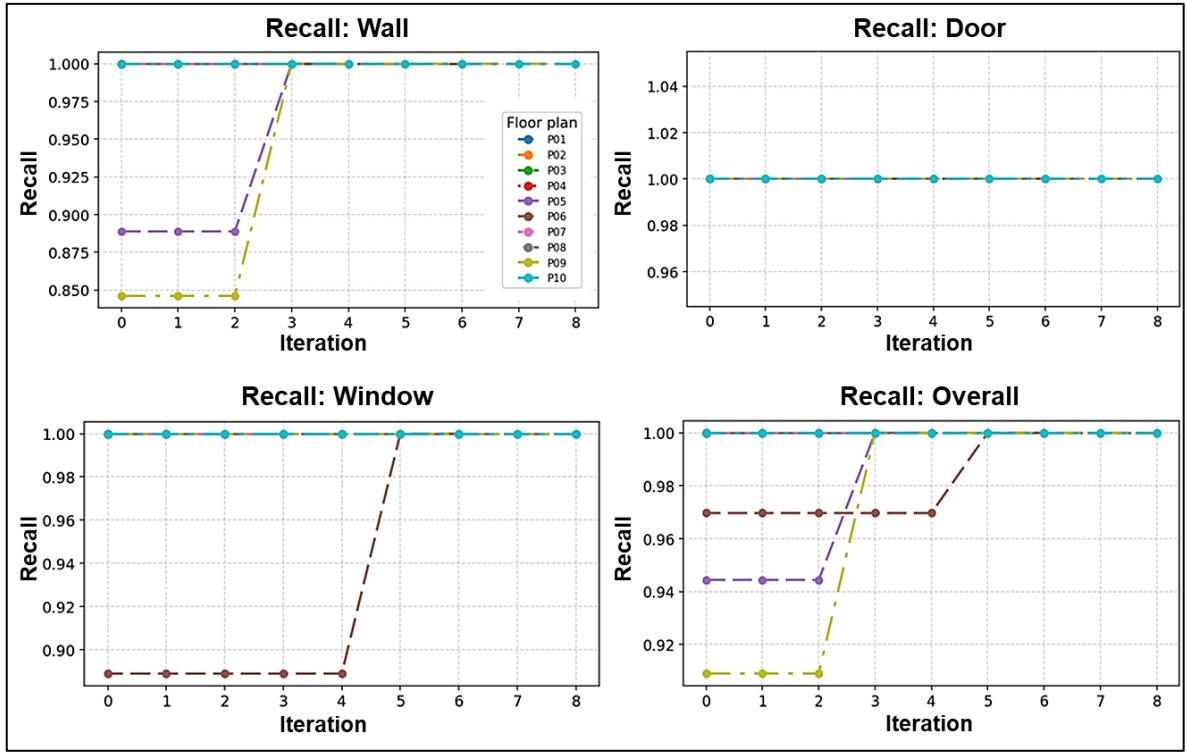

**Fig. 8.** Recall vs. human-feedback iteration by floor plan (P01–P10) and element.

Fig. 8 presents recall trajectories. Walls generally begin with a recall value of more than 0.845 and convert to 1.00 within the first three iterations. The overall recall follows this pattern, converging soon after walls. This converging of recall curves across iterations confirms that the feedback process not only eliminates false positives but also systematically recovers missed elements.

Fig. 9 synthesizes these patterns in $F1$ trajectories. The F1 score rises sharply during the first few iterations and then plateaus near 1.0, with only small gains after the third or fourth iteration. The dense cluster of dashed lines flattening at 1.0 indicates that most floor plans reach stable, near-perfect performance after just a few correction cycles. For doors, the F1 score is already 1.0 in the initial extraction, showing that openings were detected almost perfectly from the start. Walls converge quickly, while windows take slightly longer to reach the same level, as users typically refine window placement after correcting wall geometry. Overall, the F1 trend follows the walls closely, confirming that structural accuracy drives the overall detection quality.

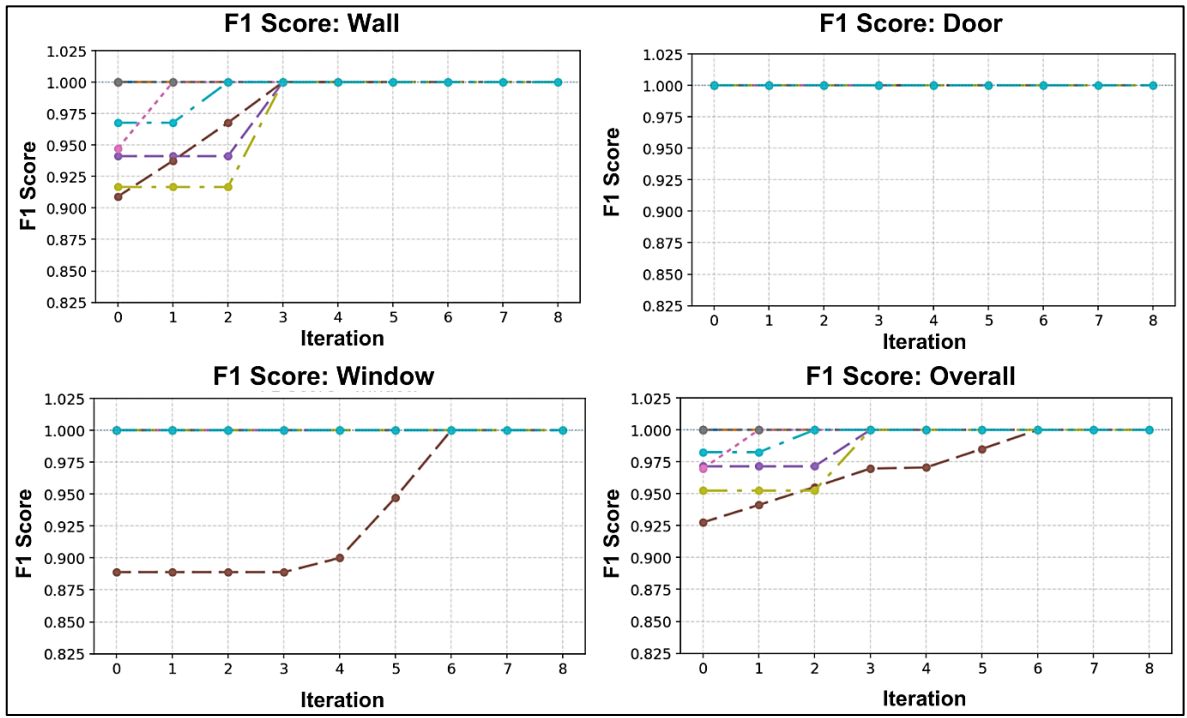

**Fig. 9.** F1 vs. human-feedback iteration by floor plan (P01–P10) and element.

It is worth noticing that in some iterations, performance metrics remain constant because the human user may focus on correcting geometric alignment or placement errors rather than detection errors, meaning that updates in precision, recall, and $F1$ depend directly on the type of correction being applied. Moreover, when a correction targets a specific element type (e.g., doors), the metrics for other elements (e.g., walls or windows) remain unchanged until those categories are explicitly addressed.

Taken together, these results demonstrate that Phase 1 layout extraction is both accurate and efficient. Structural elements such as walls are detected with high reliability from the first iteration, while more variable features, doors and windows, benefit from a limited number of human-in-the-loop refinements. Convergence across all metrics within three to four iterations validates the effectiveness of the feedback-driven multi-agent pipeline.

## 7.2. Aggregate Geometric Fidelity (Phase 1)

Geometric quality was evaluated over matched (i.e., true positive) elements using RMSE for length error and MAE for positional error (midpoint displacement). Results were presented for three categories: walls, doors, and windows. To illustrate the outcomes for each category, (i) RMSE line plots across iterations, (ii) MAE line plots across iterations, (iii) heatmaps of $\Delta$RMSE to final, and (iv) heatmaps of $\Delta$MAE to final were plotted, where $\Delta$ denotes the difference between the current value and each row's GT value.

In Fig. 10, the RMSE for wall curves was observed to start high and decrease rapidly, indicating an early correction of wall sizing. At initialization, the average wall RMSE was 2.34 ft with a range of 0.00 ft to 8.35 ft, and the average wall MAE was 0.620 ft with a range of 0.00 ft to 2.31 ft. The $\Delta$-heatmaps showed darker cells in the first iterations for several complex plans, which gradually faded to uniformly light rows, signaling convergence to each plan's GT value. By the final recorded iteration, both wall RMSE and MAE had converged to 0.00 ft.

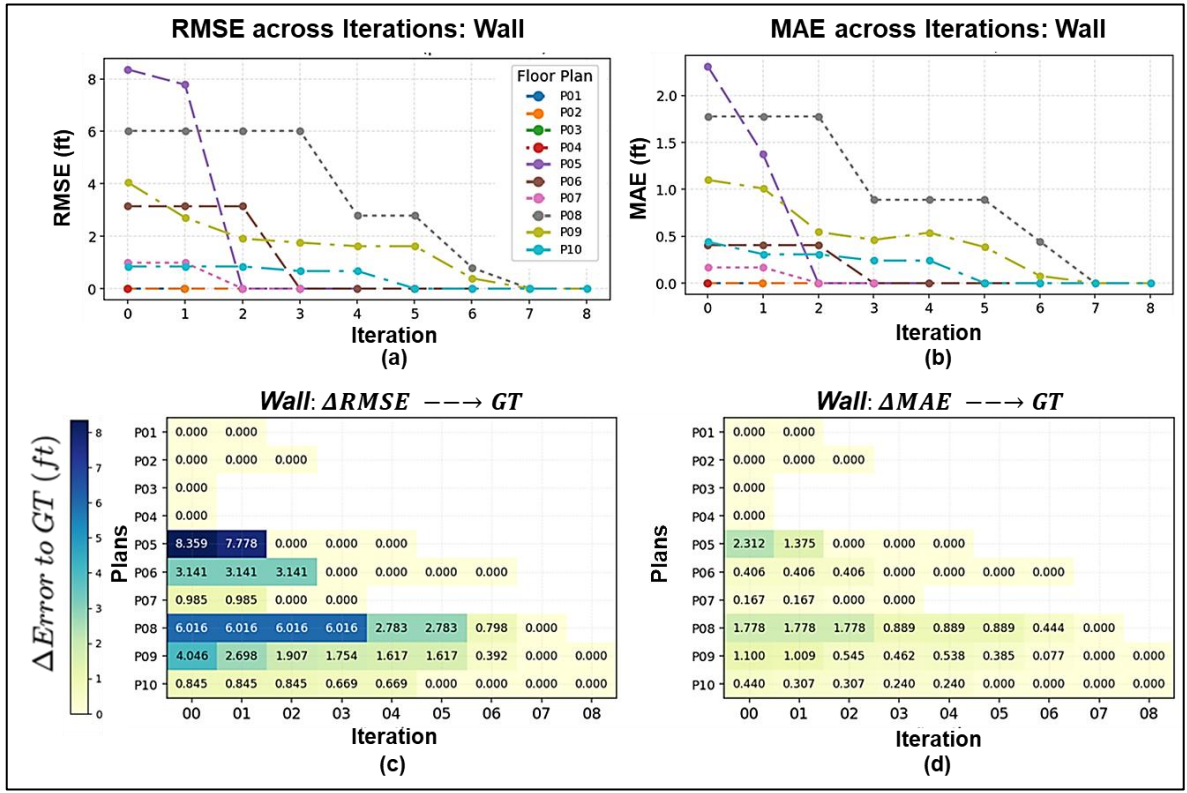

**Fig. 10.** Walls: aggregate geometric fidelity (Phase 1). (a) RMSE vs. iteration by floor plan, (b) MAE vs. iteration by floor plan, (c) ΔRMSE heatmap across plans vs. iterations, (d) ΔMAE heatmap across plans vs. iterations.

In Fig. 11, the door length error was found to be near zero for most plans, with an initial average RMSE of 0.106 ft (range: 0.00–0.65 ft). In contrast, the positional error was more pronounced, with an average MAE of 0.802 ft (range: 0.00–2.60 ft). RMSE lines remained nearly flat near zero across iterations, whereas MAE lines declined over a few steps and, in more complex cases, required one additional iteration. This pattern was consistent with darker early cells in the ΔMAE heatmap. By the final iteration, both door RMSE and MAE had converged to 0.00 ft.

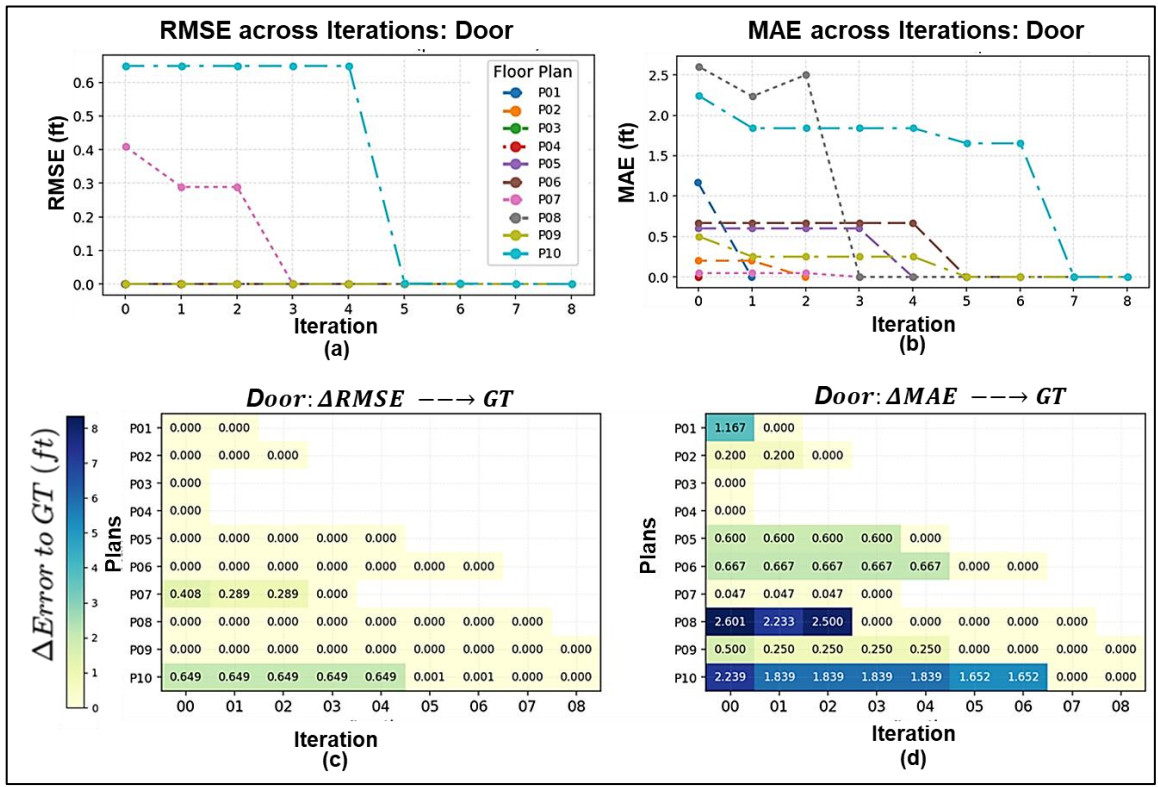

**Fig. 11.** Doors: aggregate geometric fidelity (Phase 1). (a) RMSE vs. iteration by floor plan, (b) MAE vs. iteration by floor plan, (c) ΔRMSE heatmap across plans vs. iterations, (d) ΔMAE heatmap across plans vs. iterations.

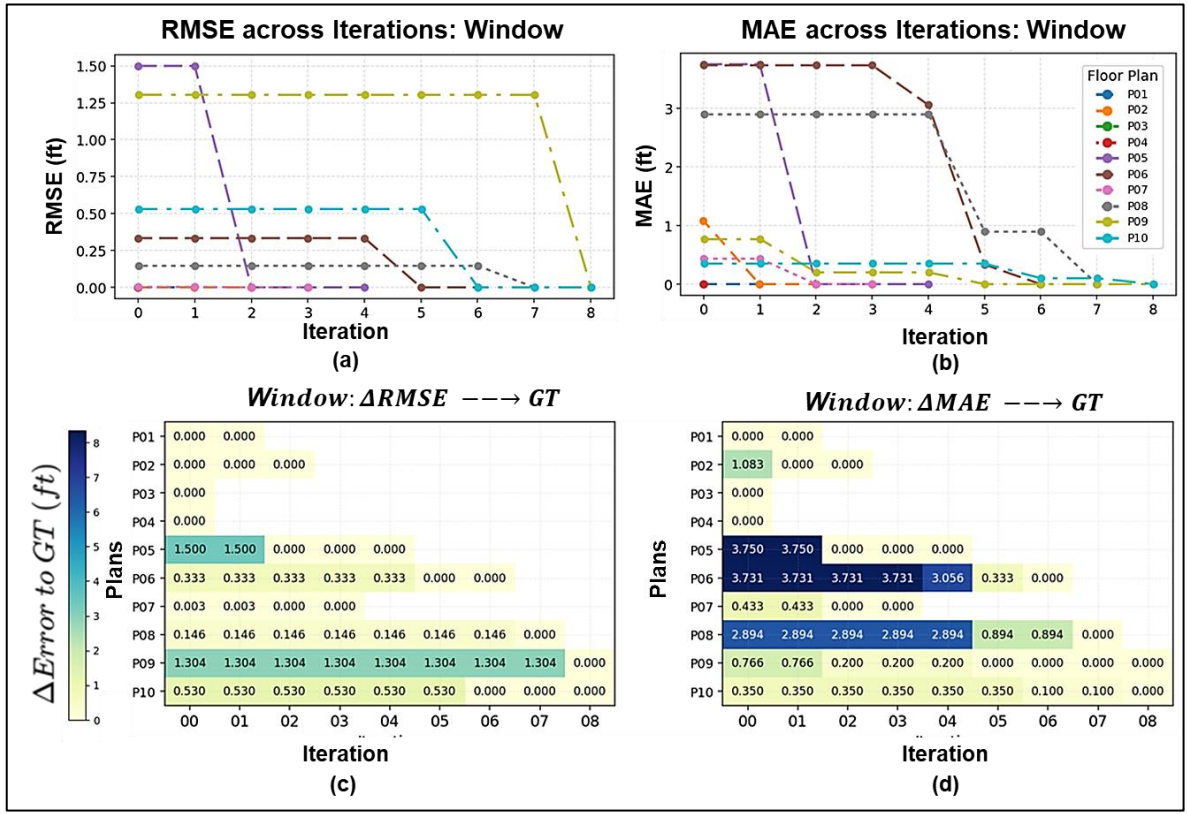

**Fig. 12.** Windows: aggregate geometric fidelity (Phase 1). (a) RMSE vs. iteration by floor plan, (b) MAE vs. iteration by floor plan, (c) ΔRMSE heatmap across plans vs. iterations, (d) ΔMAE heatmap across plans vs. iterations.

In Fig. 12, window elements were observed to have modest length deviations; however, the most significant deviation was the initial positional drift. At initialization, the average RMSE was 0.382 ft (range 0.00–1.50 ft), and the average MAE was 1.301 ft (range 0.00–3.75 ft). RMSE lines decreased rapidly, while MAE lines exhibited the steepest decline among all three categories in the early stages. The Δ-heatmaps revealed substantial early offsets that gradually diminished to lighter tones by the final iteration, with both metrics ultimately converging to 0.00 ft overall.

It should be noted that geometric fidelity metrics, similar to detection performance, may remain unchanged across certain iterations depending on the type of correction applied by the user. When human feedback is directed toward categorical errors (such as adding or removing elements), RMSE and MAE values for length and position tend to remain constant until direct geometric refinements are performed. Furthermore, when corrections are applied to a specific element type (for example, resizing walls), the metrics for other categories (such as doors or windows) remain unaffected until those elements are explicitly adjusted. This dependency indicates that improvements in geometric fidelity are strongly influenced by the correction strategy applied at each iteration.

Overall, the line plots (trajectories) and $\Delta$-heatmaps (distance-to-GT) demonstrate two Phase 1 failure modes: over- or under-sizing (RMSE) and spatial drift (MAE). Both are shown to be effectively resolved through human feedback within a limited number of iterations. Walls were initially associated with the largest length errors, but these were corrected rapidly. In contrast, windows exhibited the largest initial positional errors and were quickly aligned. Door placement occasionally required an additional iteration, but convergence was ultimately achieved. The progression to 0.00 ft for both RMSE and MAE across categories highlights the central role of human feedback in achieving geometric fidelity.

### 7.3. Validation of Phase 2

The validation of Phase 2 aimed to confirm that the BIM models created from the structured JSON layouts were both geometrically precise and semantically functional in Revit. All generated components, such as walls, doors, and windows, were instantiated as available generic families. In addition to vertical elements, a single continuous floor slab was built to cover the entire building footprint, ensuring that all enclosed rooms were properly supported.

Human validation was performed on all ten floor plans by three graduate students with prior training in BIM. As evaluators, they examined each BIM model to ensure that walls were correctly positioned and openings, such as doors and windows, were appropriately integrated into their host walls, and that connections between elements were intact without gaps or misalignments. For the floor slab, the review confirmed it completely covered the building's footprint, with no uncovered areas or unnecessary extensions.

**Table 3.** Validation checklist for floor plan 10.

| Item | Validation Criteria | Status |
|------|---------------------|--------|
| Walls | Number of walls = 15 | ✓ |
| | Spatial location is correctly aligned | ✓ |
| | Lengths consistent with ground truth | ✓ |
| | Thickness consistent | ✓ |
| Curved walls | Radius and endpoints correct | ✓ |
| | Orientation and sweep valid | ✓ |
| Doors | Number of doors = 5 | ✓ |
| | Correct wall association | ✓ |
| | Dimensions consistent | ✓ |
| Windows | Number of windows = 8 | ✓ |
| | Correct wall association | ✓ |
| | Dimensions consistent | ✓ |
| Rooms | Number of enclosed rooms = 5 | ✓ |
| | Room polygons are closed and valid | ✓ |
| Floor slab | Single continuous slab covering full footprint | ✓ |
| | No gaps or overlaps | ✓ |
| Connectivity | All wall endpoints are connected | ✓ |

In all ten cases, human validation confirmed that the generated BIM models were spatially accurate in both location and size. The single slab sufficiently covered all enclosed rooms, and all elements aligned with the intended design. These validation results align with the Phase 1 quantitative metrics, as high precision and recall indicate that all elements were detected, while low RMSE and MAE values suggest accurate geometry. Overall, the Phase 1 metrics combined with Phase 2 human validation demonstrate that the pipeline provides both numerical reliability and practical usability in BIM generation. Table 3 presents the validation checklist applied to floor plan 10.

## 8. LIMITATION

This pipeline is subject to a few limitations that highlight areas for future refinement. The current implementation does not include functions for generating architectural features such as staircases, curtains, or multi-story building components, nor does it yet support Mechanical, Electrical, and Plumbing (MEP) elements. Instead, it focuses on producing a foundational BIM model that comprises core structural components, such as walls, doors, windows, and slabs. For Manhattan-style floor plans, slabs can be created room by room using four-point boundaries, but for non-Manhattan or irregular layouts, the system generates a single continuous slab. This guarantees coverage but reduces flexibility in how floors are represented. Performance also depends on the underlying multimodal LLM (GPT-5). While generally reliable, the model occasionally hallucinates or fails. In our ten test sketches, two initial extractions had to be rerun, and one JSON-to-script conversion required repair by the correction agent ($A_6$). Because the system is not fine-tuned, its accuracy remains tied to the MLLM's stability, and outcomes may vary if the model changes. Finally, evaluation was limited to ten hand-drawn sketches, although they stress-test common challenges, they do not capture the full range of real-world floor plans. Larger-scale testing on more diverse datasets is needed. Although these limitations remain, the pipeline shows strong potential to democratize BIM creation, allowing anyone who can sketch a floor plan to generate accurate 3D models without model training or BIM expertise.

## 9. DISCUSSION

The results demonstrate that the proposed two-phase pipeline effectively combines automated layout extraction with human-in-the-loop refinement. Detection performance in Phase 1 reached high accuracy within only a few iterations of feedback. Doors and windows were identified almost perfectly in the initial extraction, while walls required more correction but converged reliably by the third or fourth iteration. This pattern illustrates the benefit of a workflow where a draft layout is produced automatically and then incrementally improved, instead of relying on a single pass of automation.

Geometric fidelity analysis highlighted two standard error types: over- or under-sizing of elements, captured by RMSE, and positional drift, captured by MAE. Both errors were most pronounced in early iterations and diminished rapidly through feedback. Heatmaps of $\Delta RMSE$ and $\Delta MAE$ confirmed that more complex cases required additional edits at the start but converged quickly to negligible error levels. These findings indicate that the pipeline is suitable for interactive use, since most corrections are front-loaded, and little effort is needed for final adjustments.

Validation of Phase 2 provided further assurance that accurate detection translates into usable BIM outputs. Graduate students verified that the generated models in Revit preserved correct geometry, element connectivity, and footprint coverage through a single continuous slab. This is important because numerical detection scores alone cannot confirm downstream usability. The alignment of Phase 1 metrics with Phase 2 validation confirms the pipeline's capacity to deliver both quantitative accuracy and practical BIM models.

## 10. CONCLUSION

This study developed a hybrid automated and interactive pipeline for converting hand-drawn floor plans into semantically consistent BIM models. The approach was evaluated across ten sketches containing both Manhattan and non-Manhattan layouts, including angled and curved walls. The findings demonstrate that the pipeline achieves high detection performance, strong geometric fidelity, and consistent usability in BIM software. Precision, recall, and F1 scores stabilized at near-perfect values within three to four iterations. RMSE and MAE errors were reduced to negligible levels through human-BIM interaction, and Phase 2 validation confirmed the production of spatially and semantically correct BIM models.

The contributions of this work are fourfold. First, it establishes a framework for directly processing noisy, unscaled sketches without reliance on CAD-like inputs or large annotated datasets. Second, it integrates multimodal reasoning with geometric extraction to handle both rectilinear and curved geometries. Third, it embeds structured feedback within 2D coordinate extraction, increasing reliability beyond one-shot automation. Fourth, it orchestrates perception, validation, and synthesis in a modular manner, reducing manual modeling effort while retaining industry-standard accuracy. These contributions demonstrate that sketch-to-BIM conversion can be made accurate, transparent, and accessible, providing a pathway to reduce barriers for both expert and non-expert users in adopting BIM.

## CREDIT AUTHORSHIP CONTRIBUTION STATEMENT

Abir Khan Ratul, contributed to conceptualization, methodology, software, data curation, formal analysis, visualization, writing original draft, and writing review and editing. Sanjay Acharjee, contributed to conceptualization, validation, data curation, software, and writing review and editing. Somin Park, contributed to conceptualization, supervision, and writing review and editing. Md Nazmus Sakib, contributed to methodology, supervision, and writing review and editing.

**DECLARATION OF COMPETING INTEREST**

The authors declare that they have no known competing financial interests or personal relationships that could have appeared to influence the work reported in this paper.

**ACKNOWLEDGMENTS**

**DATA AVAILABILITY**

The data supporting the findings of this study are available from the corresponding author upon reasonable request.

**DECLARATION OF GENERATIVE AI AND AI-ASSISTED TECHNOLOGIES IN THE MANUSCRIPT PREPARATION PROCESS**

During the preparation of this work, the authors used Grammarly and Gemini to check for plagiarism and grammar. After using these tools, the authors reviewed and edited the content as needed and take full responsibility for the content of the published article.